\def\year{2020}\relax
\documentclass[letterpaper]{article} % DO NOT CHANGE THIS
\usepackage{aaai20}  % DO NOT CHANGE THIS
\usepackage{times}  % DO NOT CHANGE THIS
\usepackage{helvet} % DO NOT CHANGE THIS
\usepackage{courier}  % DO NOT CHANGE THIS
\usepackage[hyphens]{url}  % DO NOT CHANGE THIS
\usepackage{graphicx} % DO NOT CHANGE THIS
\usepackage{graphicx}  % DO NOT CHANGE THIS
\usepackage{ragged2e}
\usepackage{enumitem}
\usepackage{color}
\definecolor{guzcolor}{rgb}{0.0,0.2,0.8}

\title{Generating Gameplay-Relevant Art Assets with Transfer Learning}
\author{Adrian Gonzalez,\textsuperscript{\rm 1} Matthew Guzdial,\textsuperscript{\rm 2} Felix Ramos\textsuperscript{\rm 1}\\ % All authors must be in the same font size and format. Use \Large and \textbf to achieve this result when breaking a line
\textsuperscript{\rm 1}Department of Computer Science, Cinvestav IPN, Unidad Guadalajara, M\'exico\\ %If you have multiple authors and multiple affiliations
\textsuperscript{\rm 2}Computing Science Department, University of Alberta\\
adrian.glez@cinvestav.mx, guzdial@ualberta.ca, felix.ramos@cinvestav.mx% email address must be in roman text type, not monospace or sans serif
}
 
\begin{document}

\maketitle
\begin{abstract}

In game development, designing compelling visual assets that convey gameplay-relevant features requires time and experience.
Recent image generation methods that create high-quality content could reduce development costs, but these approaches do not consider game mechanics. 
We propose a Convolutional Variational Autoencoder (CVAE) system to modify and generate new game visuals based on their gameplay relevance.
We test this approach with Pok\'emon sprites and Pok\'emon type information, since types are one of the game's core mechanics and they directly impact the game's visuals.
Our experimental results indicate that adopting a transfer learning approach can help to improve visual quality and stability over unseen data.

\end{abstract}

\section{Introduction}

Game development is a demanding task. 
Gameplay systems generally include numerous elements to make them stand out from similar titles, as well as to provide variety and balance. 
On the other hand, designing compelling visual assets that quickly and consistently convey those gameplay-relevant features (play-style, difficulty, weaknesses, etc.) is not trivial, especially while striving to preserve project-wide artistic cohesion.
This is also an important consideration when creating variations on existing content, such as characters' alternative appearances or \textit{skins}, enemy sub-classes (e.g., Mario's Dry Bones are visual and mechanical variations of the Koopas), or player customization systems. Most of these processes are iterative and time demanding, further increasing development costs \cite{Reboucas2019pixelart}.

Automating the visual design process could help to improve asset quality and reduce development time.
Recent general-purpose deep-learning models for image creation provide high-quality results; however, these approaches are limited to particular tasks with large training sets, such as face, character, or landscape generation \cite{isola2016imagetoimage,karras2018,artbreeder2020}.
Outside of fully autonomous generation, some approaches identify latent vectors to grant users the ability to explore the possibilities of a model's learned latent space \cite{burgess2018betavae,voynov2020interpretablelatentspace}.
We identify two main issues with both of these approaches: they require large amounts of data and their controllability is not influenced by gameplay-relevant aspects like mechanics.

To study how we can generate images that relay gameplay-relevant information, we decided to work with images from the Pok\'emon series \cite{pokemon2020}.
Pok\'emon games have clearly defined gameplay elements that are present in their art style (the \textit{type information}), which help to communicate each Pok\'emon's strengths and weaknesses to players \cite{liapis2018recomposingtp}. 
The main Pok\'emon titles are turn-based role-playing videogames in which players make their companions --Pok\'emon-- battle. Understanding each type's weaknesses and resistances is crucial for victory.

In this article, we present a Variational Autoencoder (VAE) for Pok\'emon type swapping, which modifies the inputs' visual designs to transmit user-defined types in a controllable manner.
To combat the problem of low training data, we adopted a transfer learning approach.
Our system's intended behavior would change a specified Pok\'emon's appearance to convey a given target type (such as \textit{fire} or \textit{water}), according to the visual attributes commonly exhibited by Pok\'emon of that target type. 
For instance, if the yellow-colored electric-type Pok\'emon \textit{Pikachu} were changed to fire-type, its colors would shift towards red. 

\section{Related Work}

In this section, we briefly discuss computational approaches to automated or assisted visual design generation.
We also present an overview of mainstream generative models and their applications to the production of visual game assets, and how those works relate to our proposed approach.

\subsection{Procedural Content Generation}

Procedural content generation or PCG refers to the creation of game content using algorithms with limited or indirect user input \cite{shaker2016}. 
More closely related to our proposal's objective, there is Visual PCG \cite{guzdial2017visualpcg}, which involves the generation of visual components for games, and PCG via machine learning (PCGML) \cite{summerville2018pcgml}.
We present some PCG-based works that create visual game elements and how a particular game's mechanics affect their creation processes.

Pollite \cite{guzdial2017visualpcg} is an artificial abstract artist based on a convolutional neural network (CNN) that learns to associate features, like shapes and colors, to emotions, from tagged real-world pictures. It can create and modify images to express feelings such as anger or joy.
Both their work and our proposal involve concepts to alter visuals (emotions and type information, respectively). 
However, adopting an approach similar to Pollite's would require tagging real-world scenes with gameplay elements that, in many cases, would demand manually annotating them or creating a system to do it instead, thus reducing the expected development benefits.

The evolutionary algorithm developed by \cite{liapis2018recomposingtp} modifies Pok\'emon sprites' colors based on type information. It uses the associations given between color palettes and the different Pok\'emon types, e.g., \textit{fire} type is related to red tones. Although their approach and ours aim to assist artists in their tasks, their system evolves a sprite's color palette and then assigns them a type, in contrast, we propose to define a type (or types) and then change the Pok\'emon's colors and shape (and even textures) to fit the new type information. This increases our model's expressiveness, since it is not limited to palette swaps, and allows its users to make specific requests, such as a \textit{fire}-type \textit{Pikachu}.

\subsection{Deep Generative Models}

In this subsection, we mention works based on two deep-learning architectures applied to image generation: Variational Autoencoders (VAEs) \cite{kingma2013vae} and Generative Adversarial Networks (GANs) \cite{goodfellow2014}. 
A VAE consists of two networks: first, an encoder that generates a mean and a variance of a Gaussian distribution per latent space dimension, then the inputs' latent representations are obtained by sampling from such distributions, and second, a decoder that reconstructs those representations back to the input data space \cite{kingma2013vae,larsen2015autoencodinglearnedsimilarity,pihlgren2020improvingembeddings}. 

Works that use VAEs for visual design in games are uncommon. Nonetheless, VAEs have been employed for image and texture synthesis \cite{chandra2017texturesynthesis,kingma2019introductionvaes}, and level generation \cite{guzdial2018explainabledesignpatterns,snodgrass2020multidomain}.
We decided to use a VAE since points sampled from the latent space near a known input tend to resemble it, which is useful to create variations of existing content. However, as stated in \cite{kingma2019introductionvaes}, generative VAEs are known to produce blurry results. Therefore, we consider exploring GANs as future work.

In \cite{Reboucas2019pixelart}, the authors proposed a deep-learning asset generation system for pixel art sprites for a 2D fighting game using line art sketches. Their tool, which is built upon the pix2pix architecture \cite{isola2016imagetoimage}, produces semi-final sprites that must be fine-tuned by a human artist, therefore reducing the production time for each image. 
Unlike our proposal, they do not include gameplay-related information in their model.

Artbreeder \cite{artbreeder2020}, a tool based on a deep convolutional GAN (DCGAN), allows its users to manipulate numerous parameters to adjust the creation of images such as human faces, landscapes, and characters. 
Another relevant DCGAN model is presented in \cite{horsley2017spritedcgan}, which generates sprites of faces, characters, and creatures using small amounts of training data. The system developed by \cite{jin2017automaticanimefaces} permits its users to create anime faces using a GAN architecture, by providing parameters, such as hair and eye color and style, to control the image generation process. 
However, none of these models explicitly consider gameplay-specific features to control the generation process.

A notable related work is \textit{pokemon2pokemon} \cite{wong2019pokemon2pokemon}, which uses CycleGAN \cite{zhu2017unpairedit} to modify the color of Pok\'emon images given a type and shows positive results. However, it does not modify the Pok\'emon's shape, which might be due to CycleGAN's difficulties when handling geometric changes.

\subsection{Pok\'emon}

The main Pok\'emon titles are turn-based role-playing games in which humans command creatures named Pok\'emon during one-on-one or two-on-two battles. 
Types are a core mechanic and there exist 18 types: \textit{Bug}, \textit{Dark}, \textit{Dragon}, \textit{Electric}, \textit{Fairy}, \textit{Fighting}, \textit{Fire}, \textit{Flying}, \textit{Ghost}, \textit{Grass}, \textit{Ground}, \textit{Ice}, \textit{Normal}, \textit{Poison}, \textit{Psychic}, \textit{Rock}, \textit{Steel}, and \textit{Water}. 
Each type possesses weaknesses and resistances to \textit{attacks} from other types. 
Every Pok\'emon has one or two types and four attacks (each with its own type). 
An attack's damage depends on the attacked Pok\'emon's types weaknesses and resistances.
Pok\'emon who use an attack that matches their type gain increased effect.
This makes understanding types crucial to win. 
For instance, fire-type Pok\'emon are weak against water-type attacks, but resistant to grass-type ones.
Therefore, conveying the types of a Pok\'emon through its design is crucial, especially in the earlier games of the series in which players were not shown the types of a Pok\'emon unless they owned it.

\section{System Overview}

\begin{figure*}[h]
\centering
\includegraphics[width=0.9\textwidth]{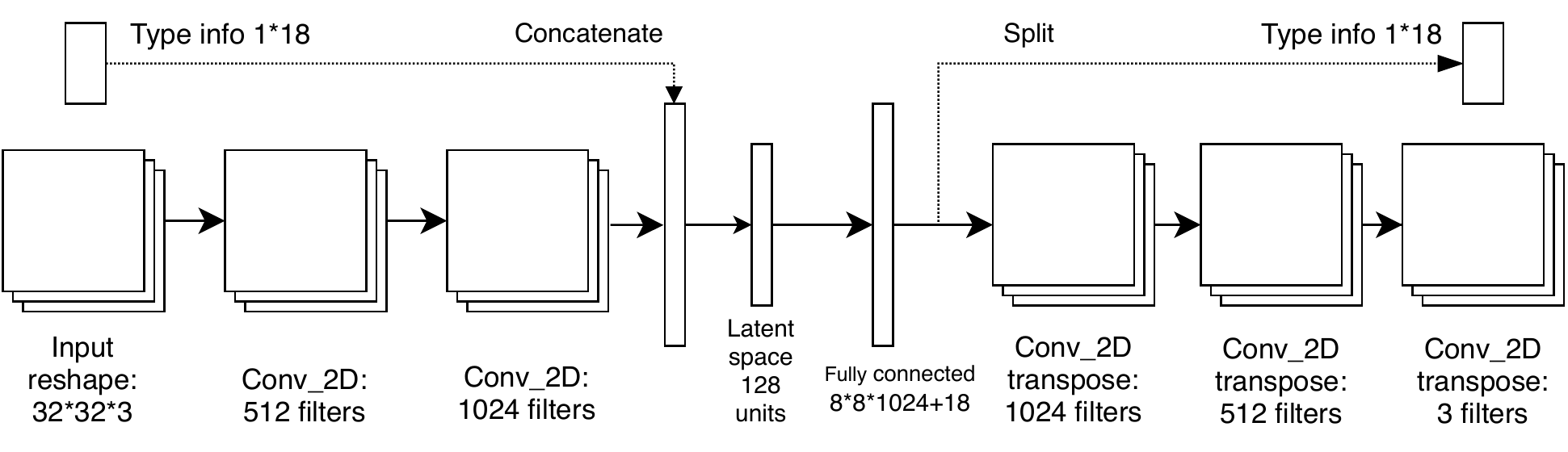} % Reduce the figure size so that it is slightly narrower than the column. Don't use precise values for figure width.This setup will avoid overfull boxes. 
\caption{Architecture of the proposed convolutional variational autoencoder (CVAE). Type information is represented as a one-hot-encoded vector of size 18. The type information and the last convolution's result are passed to the latent space's fully connected layers.}
\label{fig:vae_architecture}
\end{figure*}

Our objective is a system that allows its users to create variations on existing Pok\'emon. The users select a Pok\'emon design and one or two types, then the image is modified to make it show distinctive features of the given types.

To achieve this, we employ a convolutional VAE. Our process to train our VAE is as follows.
First, we collect a set of Pok\'emon images and their type information for training. 
Second, given the lack of data, we use the Anime Face Dataset (AFD for short) \cite{churchill2019animefacedataset} as a source dataset for a transfer learning approach. 
Given our final goal of controllability through Pok\'emon type information, we assign type labels to the AFD's images based on how similar they are to each type's Pok\'emon designs.
Then, we train the VAE on the now-labeled dataset.
Finally, we transfer the learned weights from the anime samples and fine-tune them via training on the Pok\'emon images.

\subsection{Dataset Collection}

The Pok\'emon images and type information were retrieved from \cite{kvpratama2017pokemondataset} and \cite{subbiah2018pokemontypes}, respectively, and updated with resources from \cite{serebii2020pokemonwebpage}.
Some elements, such as the \textit{Pikachu} variations, were omitted to avoid over-representing features in the set. 
Our final set contained 974 Pok\'emon and is available online\footnote{\url{https://github.com/EtreSerBe/PokeAE}}.
Additionally, we used the Pok\'emon \textit{regional variants}, which are variations of Pok\'emon but with different types and slightly distinct designs (to convey their modified types), to build a special test dataset. This set provides us with useful comparison data for our model.
Since a Pok\'emon can have one or two types, and there exist 18 Pok\'emon types, we handled type information as one-hot-encoded vectors of size 18, and used 0.5 in two positions for Pok\'emon with two types.
We experimented with a one-hot encoding for types, but found it less effective for our needs.

All images were resized to 32*32 pixels using bicubic filtering, the same size as in \cite{krizhevsky09cifar10}, and converted to the Hue, Saturation, Value (HSV) format, like in \cite{liapis2018recomposingtp,lim2016aestheticavatars}. 
Since dark-colored Pok\'emon were showing poor results, we opted to use four different background colors for each sample: black, white, and two random noise backgrounds (for training samples only). As in \cite{Reboucas2019pixelart}, we only used horizontal flips for data augmentation; thus, we had eight images per Pok\'emon, for a total of 7204 instances in our dataset.

Given that the results obtained with the Pok\'emon images were not sharp nor detailed enough in initial tests, we adopted a transfer learning approach using a dataset that shared some visual traits with our target domain. Pok\'emon designs resemble some Japanese manga and anime styles; hence, we decided to work with the Anime Face Dataset (AFD) \cite{churchill2019animefacedataset}, which contains about 63,000 illustrations of anime-style character faces. We augmented these by flipping horizontally as well.

\subsection{Transfer Learning Process}

The AFD does not possess type information, which is crucial for our intended system. To provide the AFD with the types required for the transfer learning process, and to ensure that the distribution of the types in both datasets was equal, we did the following: first, we obtained the mean HSV value for each of the 18 Pok\'emon types, considering only non-background pixels in each image. Second, for every element in the anime set, we calculated the mean HSV value of its pixels and computed its mean squared distance with respect to each of the types' average HSV values. 
Third, we used these distances as preferences (the lowest one being the most preferred), and then employed the Gale-Shapley algorithm \cite{galeshapley1962} to assign the types. 
In the current implementation, each image was given only one type.

We decided to use a VAE architecture because they can reproduce given inputs with slight modifications. This behavior is beneficial to our goal since we want the modified designs to be recognizable as variations of the original one; thus, some of the source's characteristics must be preserved, and the changes made should be enough to convey the new type information. 

The proposed convolutional VAE (CVAE) model is shown in Figure \ref{fig:vae_architecture}. It is similar to the CVAE shown in \cite{tensorflow2020cvae} but adapted for HSV format, and the type information is handled like the \textit{level design pattern labels} used in \cite{guzdial2018explainabledesignpatterns}.
Our model receives the images in HSV format plus the vector of type information. The model's encoder consists of two convolution operations (512 and 1024 filters respectively) with 2x2 filters and 2x2 strides (instead of \textit{max-pooling} \cite{horsley2017spritedcgan}). The filter sizes are small because the resulting images lacked detailed features (cloudiness). The second convolution's output and the given type information are fed to the two fully connected layers with 128 units each for the latent space (mean and standard deviation).
The decoder takes a vector of size 128 as input, which is passed to a fully connected layer of size 8*8*1024+18; after that, we split the last 18 values to reconstruct the type information. The remainder is passed through two deconvolutions \cite{zeiler2010deconvolutionalnetworks} with 1024 and 512 filters, respectively. Finally, it passes to another deconvolution with only three filters for the output image's HSV values. We used the Adam optimizer, and our loss function was the reduced mean of the sum of the cross-entropy and the Kullback-Leibler divergence, as proposed in \cite{kingma2013vae}. All layers' activation functions were leaky relu, except for the latent space, which was linear, and the output's activation, which used relu.

The model was trained first on the AFD with the added type information.
The initial training stage consisted of 10 epochs with a learning rate of 0.0001 and a batch size of 128. In later stages, we fine-tuned the model by decreasing the learning rate to 0.00001 and training for 50 more epochs. 
Then, we fine-tuned the model on the Pok\'emon dataset, with a learning rate of 0.0001. 
We trained with a batch size of 256 for three rounds of 100 epochs each, with learning rates of 0.00005, 0.00002, and 0.00001, respectively.
The low learning rates were used so the convolutional filters learned from the anime samples did not change abruptly, as they would lose the benefits of transfer learning. 

We randomly split the Pok\'emon dataset into 6616 training instances and 588 test instances, with 827 and 147 different Pok\'emon, respectively. Our type-labeled AFD contains 125,130 training images and only 2,000 for testing.
To be used as a baseline for comparison, we trained another instance of our model using only the Pok\'emon data. 
It was trained for 50 epochs with a learning rate of 0.0001 and a batch size of 128. 
Later, we fine-tuned it by training for ten rounds of 50 epochs each, with a batch size of 256.

\section{Evaluation}

The evaluation is focused on measuring our system's outputs' visual quality (since the Pok\'emon must be detailed enough to be recognized as the one given as input), and controllability based on the type information set by the user. 
We performed three evaluation tasks.
To determine the generated images' quality, we compared them to the input images provided, over both test and train sets.
We used two comparison metrics: Mean Squared Error (MSE) in the RGB images and Structural Similarity Index (SSIM) \cite{wang2004ssim}, in YUV format, with filter size=11, filter sigma=1.5, k1=0.01, and k2=0.03.

\begin{figure}
\centering
\includegraphics[width=0.95\columnwidth]{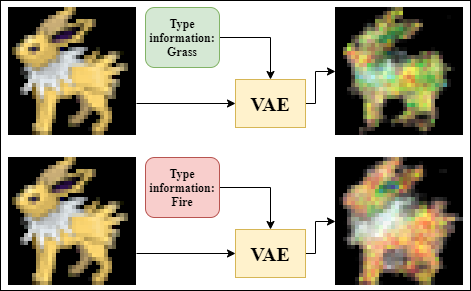} % Reduce the figure size so that it is slightly narrower than the column. Don't use precise values for figure width.This setup will avoid overfull boxes. 
\caption{Type swap task example. The electric-type Pok\'emon on the left is modified to show \textit{grass} (above) and \textit{fire} (below) types. The one above shows greener tones that are common in grass-type Pok\'emon, while the one below has bright red colors like many fire-type Pok\'emon.}
\label{fig:type_swap_overview}
\end{figure}

On the other hand, the procedure to evaluate the controllability or \textit{type-swap task} consisted of setting the type for every sample to a single target type and passing them to the system, instead of their original types. This process is shown in Figure \ref{fig:type_swap_overview}.
Note that only one type was used since the anime dataset samples were only assigned one type each. We tested this with four types: \textit{fire}, \textit{grass}, \textit{water}, and \textit{fairy}.
We used the first three since most of the Pok\'emon of those types are red, green, and blue respectively.
We used the fairy type because it was the second most preferred type during the anime faces type assignation before applying the Gale-Shapley algorithm. 

Additionally, we performed a third evaluation that involves the previous two and a special \textit{regional variants} test dataset. 
This dataset was composed of in-game variations of existing Pok\'emon where they possess different types and designs (to convey their modified types). The \textit{original-to-regional} task consisted of comparing the visual similarity between the regional variants' images and our system's output after type-swapping their non-regional versions to the variants' types. 

\section{Results}

\begin{figure}
\centering
\includegraphics[width=0.95\columnwidth]{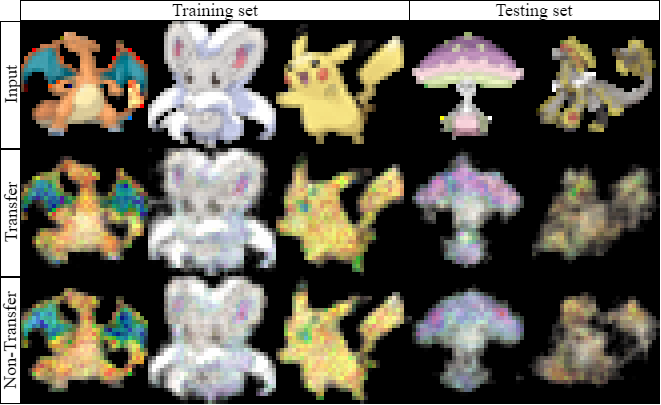} % Reduce the figure size so that it is slightly narrower than the column. Don't use precise values for figure width.This setup will avoid overfull boxes. 
\caption{Reconstruction accuracy comparison. Results over training data are of similar quality, but, as the SSIM comparison indicates, the images are structurally more accurate when using the transfer approach.}
\label{fig:accuracy_test}
\end{figure}

The Pok\'emon reconstruction visual quality scores are shown in Tables \ref{table:mse_results} and \ref{table:ssim_results}. 
For the MSE results in Table \ref{table:mse_results} lower is better, and the transfer learning model consistently outperforms the non-transfer one.
For the SSIM results in Table \ref{table:ssim_results} higher is better, which indicates at least a 1\% increase in visual similarity to the inputs when using the transfer learning approach, even though the two datasets are vastly different.
Note that for all of this article's figures all images shown, including the Pok\'emon inputs, were resized from 32x32 pixels to 128x128 using the nearest neighbor method.
An example of both systems' outputs is presented in Figure \ref{fig:accuracy_test}. Both models present blurry outcomes, a known drawback of autoencoders, which will be improved in future work.
 
\begin{table}
\begin{center}
\begin{tabular}{|c|c|c|c|} 
\hline
Model version & Test & Train & Test and train\\ 
\hline
Transfer learning & \textbf{0.03216} & \textbf{0.01409} & \textbf{0.01692} \\ 
\hline
Non-transfer & 0.03349 & 0.01438 & 0.01733 \\ 
\hline
\end{tabular}
\end{center}
\caption{Image comparison between actual Pok\'emon and reconstructed Pok\'emon images using Mean Squared Error.}
\label{table:mse_results}
\end{table}

\begin{table}
\begin{center}
\begin{tabular}{|c|c|c|c|} 
\hline
Model version & Test & Train & Test and train\\ 
\hline
Transfer learning & \textbf{0.3940} & \textbf{0.6496} & \textbf{0.6109} \\ 
\hline
Non-transfer & 0.38361 & 0.6268 & 0.5895 \\ 
\hline
\end{tabular}
\end{center}
\caption{Image comparison values using the Structural Similarity Index (SSIM).}
\label{table:ssim_results}
\end{table}
%A table or tables or figures that report the results. In the text walk a reader through these results.

\begin{table}
\begin{center}
\begin{tabular}{|c|c|c|} 
\hline
Model version & MSE & SSIM \\ 
\hline
Transfer learning & 0.1000 & \textbf{0.1704} \\ 
\hline
Non-transfer & \textbf{0.0952} & 0.1633 \\ 
\hline
\end{tabular}
\end{center}
\caption{\textit{Original-to-regional} visual quality task results. The transfer learning version presents slightly less pixel-wise errors but higher structural similarity with the regional variants.}
\label{table:original_to_regional_results}
\end{table}

\begin{figure}
\centering
\includegraphics[width=0.95\columnwidth]{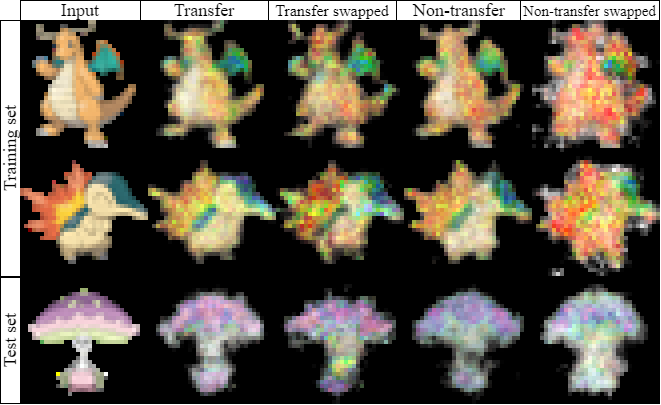} % Reduce the figure size so that it is slightly narrower than the column. Don't use precise values for figure width.This setup will avoid overfull boxes. 
\caption{\textit{Fire} type swap task results. The non-transfer model generates more evident visual changes but its effects are uncontrollable. Their original types were: \textit{dragon-flying}, \textit{fire}, and \textit{grass-fairy}.}
\label{fig:fire_swap}
\end{figure}

\begin{figure}
\centering
\includegraphics[width=0.95\columnwidth]{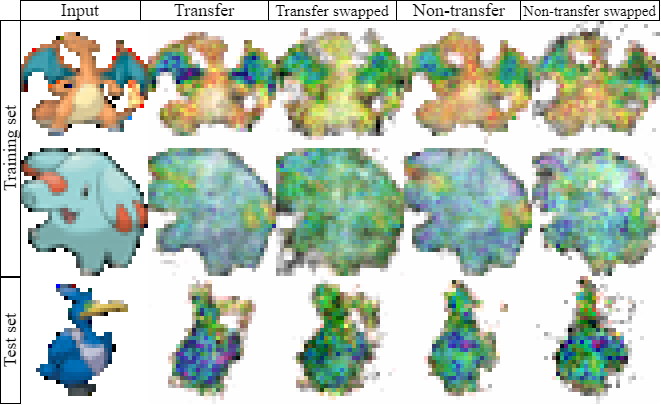} % Reduce the figure size so that it is slightly narrower than the column. Don't use precise values for figure width.This setup will avoid overfull boxes. 
\caption{\textit{Grass} type swap task results using white background samples. Both models had troubles with that background during type swap tasks. Their original types were: \textit{fire-flying}, \textit{ground}, and \textit{water-flying}.}
\label{fig:grass_swap}
\end{figure}

\begin{figure}
\centering
\includegraphics[width=0.95\columnwidth]{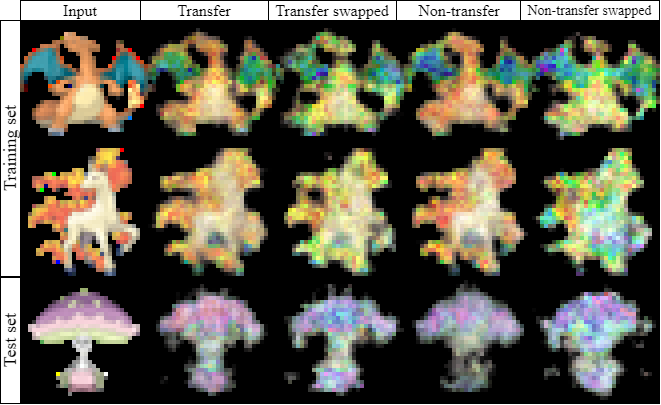} % Reduce the figure size so that it is slightly narrower than the column. Don't use precise values for figure width.This setup will avoid overfull boxes. 
\caption{\textit{Water} type swap task results. Their original types were: \textit{fire-flying}, \textit{fire}, and \textit{grass-fairy}.}
\label{fig:water_swap}
\end{figure}

Initial type-swap task outcomes were barely distinct from the inputs. To make them more evident, we increased the type vector's magnitude from 1 to 20. Representative results of the type-swap task are shown in Figures \ref{fig:fire_swap}, \ref{fig:grass_swap}, and \ref{fig:water_swap}, but the ones for fairy-type were omitted because no consistent changes were noticeable. The lack of changes may have been caused by the large number of AFD's images that preferred the fairy type but were reassigned to another one by the Gale-Shapley algorithm (13717, which is 10.88\% versus the final 4.05\% or 5109). This reassignment likely caused several images' fairy-like features to be ignored.

In Figure \ref{fig:fire_swap}, the non-transfer model generates more evident visual changes. However, its effects can be uncontrollable and make it difficult to recognize from the original, which might be caused by the amount of variety of samples. Both models' results showed the expected red colors that most fire Pok\'emon possess, but the transfer learning one also presented some unexpected vivid green tones, likely due to how the types were assigned to the AFD's images.

In Figure \ref{fig:grass_swap}, both models presented noise on the white background during type swap experiments, especially with type vectors of large magnitude. The transfer learning model's results present the expected green coloring. On the other hand, the baseline model generated very noisy images. Several AFD's samples have white background, which might explain why the transfer approach had less isolated pixels in this experiment.

Finally, in Figure \ref{fig:water_swap}, the non-transfer model's results present considerable changes, but also noticeable artifacts. The transfer model's outputs over testing data were slightly more blue, but did not show the expected blue tones when using red or yellow Pok\'emon. 
This problem could be caused by the small number of AFD's images labeled as water-type before the Gale-Shapley algorithm was used (about 2000). That is less than 2\% of the samples, compared to the 11.5\% water Pok\'emon, therefore, that difference was compensated with anime faces that might not have been suitable as water-type.
While the non-transfer version of the model shows more evident changes in this task, it also presents considerably more noise, both inside and outside of the Pok\'emon, which is not desirable in the final content. 
Therefore, with our current results, we argue that pursuing a transfer learning approach is more suitable if the generated content is aimed towards the general public, especially when the training data is scarce. 
However, we acknowledge that both models' results over testing data have plenty of room for improvement.

\begin{figure}
\centering
\includegraphics[width=0.95\columnwidth]{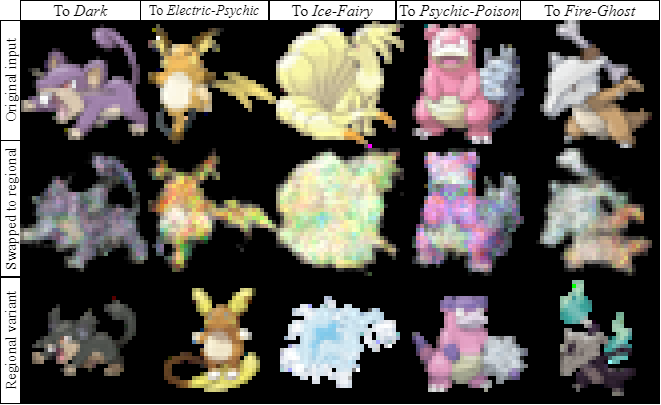} % Reduce the figure size so that it is slightly narrower than the column. Don't use precise values for figure width.This setup will avoid overfull boxes. 
\caption{\textit{Original-to-regional} task representative results. The bottom row elements are existing type-swapped versions of the Pok\'emon in the first row. The middle row ones are our system's proposed type-swapped designs. The target types are above each column. The first column shows darker tones common in dark-type Pok\'emon, and the third one seems more white (similar to most ice-types). Their original types are, from left to right, \textit{normal}, \textit{electric}, \textit{fire}, \textit{psychic-water}, and \textit{ground}.}
\label{fig:original_to_regional_sample}
\end{figure}

\begin{figure}
\centering
\includegraphics[width=0.95\columnwidth]{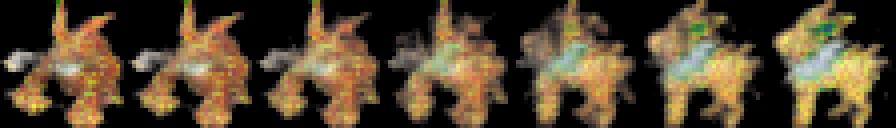} % Reduce the figure size so that it is slightly narrower than the column. Don't use precise values for figure width.This setup will avoid overfull boxes. 
\caption{Interpolation between two different Pok\'emon's latent representations. The \textit{fire}-type Pok\'emon on the left changes towards the \textit{electric}-type one on the right. This is useful as we could explore which dimensions affect features related to each type, to improve the type swap results.}
\label{fig:interpolation}
\end{figure}

For the original-to-regional task, the visual quality comparison results are shown in Table \ref{table:original_to_regional_results}. The scores are considerably lower than for the reconstruction task, in part because the regional variants not only change colors but also size, shape and pose. We also present some positive visual results in Figure \ref{fig:original_to_regional_sample}. 
We note that changing the fourth Pok\'emon (\textit{Slowbro}) from \textit{water-psychic} to \textit{poison-psychic} introduced the same purple splotches seen in the real regional variant.
We illustrate how our VAE interpolates between two different Pok\'emon in Figure \ref{fig:interpolation}.

\section{Limitations and Future Work}

We identify three crucial aspects to improve our proposed system. 
First, we consider that using a different method to assign the type information to the Anime Face Dataset could lead to improvements for the type-based controllability of the transfer model. 
This is because one HSV tuple for each type cannot hold information about spatial or structural features (for instance, most Flying-type Pok\'emon have wings). 
Moreover, the Saturation and Value channels in the type were confined to small ranges [0.22, 0.37] and [0.51, 0.69] (in contrast, the Hue values ranged between [0.19, 0.51]), which resulted in undesired type overlaps.
Additionally, several images from the AFD have white background, which causes the type assignment process to favor types with brighter palettes such as \textit{fairy} and \textit{fire}.

Second, the current network architecture used is simple. We argue that exploring other alternatives, such as using image patches, discriminator modules, or moving to different architectures (such as GANs), could lead to better image quality and controllability results. Third, we require evaluating how human artists respond to this kind of system, which will provide us with crucial feedback about future work directions.

On a more distant horizon, we also consider the exciting possibility that this kind of approach could be applied in games. For instance, it could be used to automatically generate visual indicators for Pok\'emon that are affected with special status effects, such as burned, frozen, or poisoned, which are currently only shown as a written text. 
Visual indicators like these could help to enhance a player's understanding of the game's mechanics.

\section{Conclusions}

We propose a Convolutional Variational Autoencoder (CVAE) system to modify Pok\'emon sprites according to a target Pok\'emon type. Our experimental results indicate that adopting a transfer learning approach, using a type-labeled version of the Anime Face Dataset, can help to improve visual quality and stability over unseen data, despite the considerable differences between both domains. While the presented models' outcomes might be usable during very early stages of the design process, their quality and controllability are not yet suitable for game development beyond that point. However, we expect that this problem will diminish in future versions of the system.

\section*{Acknowledgements}

This work was supported by CONACYT through the doctoral scholarship number CVU-777880. We acknowledge the support of the Natural Sciences and Engineering Research Council of Canada (NSERC) and the Alberta Machine Intelligence Institute (Amii). 

\bibliographystyle{aaai.bst}
\bibliography{main.bib}

\end{document}